\documentclass[conference]{IEEEtran}
\bstctlcite{IEEEexample:BSTcontrol}
\ifCLASSINFOpdf

\else

\fi

\usepackage{graphicx}
\usepackage{subcaption}
\usepackage{float}

\usepackage{amsmath,amssymb,epsfig}

\begin{document}
	
	\title{Chaotic Time Series Prediction using Spatio-Temporal RBF Neural Networks}
    \author{\IEEEauthorblockN{Alishba Sadiq\IEEEauthorrefmark{1}, Muhammad Sohail Ibrahim\IEEEauthorrefmark{2}, Muhammad Usman\IEEEauthorrefmark{2}, Muhammad Zubair\IEEEauthorrefmark{2} and Shujaat Khan\IEEEauthorrefmark{2}}
		\IEEEauthorblockA{\\
			\IEEEauthorrefmark{1}College of Engineering, Karachi Institute of Economics and Technology, Karachi 75190, Pakistan.\\
			Email: alishba.sadiq@pafkiet.edu.pk\\
			\IEEEauthorrefmark{2}Faculty of Engineering Science and Technology (FEST), Iqra University, Pakistan.\\
			Email: \{msohail, musman, zubair, shujaat\}@iqra.edu.pk}}
	\maketitle
	
	
\begin{abstract}
Due to the dynamic nature, chaotic time series are difficult predict. In conventional signal processing approaches signals are treated either in time or in space domain only. Spatio-temporal analysis of signal provides more advantages over conventional uni-dimensional approaches by harnessing the information from both the temporal and spatial domains. Herein, we propose an spatio-temporal extension of RBF neural networks for the prediction of chaotic time series. The proposed algorithm utilizes the concept of time-space orthogonality and separately deals with the temporal dynamics and spatial non-linearity(complexity) of the chaotic series. The proposed RBF architecture is explored for the prediction of Mackey-Glass time series and results are compared with the standard RBF. The spatio-temporal RBF is shown to out perform the standard RBFNN by achieving significantly reduced estimation error. 
\end{abstract}
	
	\IEEEpeerreviewmaketitle

\section{Introduction}
\label{intro}
Cybernetics and neural learning systems are becoming essential part of the society. They offer solution to several practical problems that occur in different areas ranging from medical sciences \cite{RAFP_Pred, ECMSRC, DNN_US} to control of complicated industrial processes \cite{industry}. With the advent of high power computing systems and deep network architectures enormous success has been achieved in diverse applications \cite{DNN_US, article5}. However, deep learning typically demands very large amount of data  to extract the useful information from training samples. The need for high computational power and sufficiently large data-sets make implementation of deep architecture unrealistic for many simple applications. Whereas, the classical neural networks still provide reasonable performance with low computational complexity and small training data-set.

Neural networks (NNs) can easily model non-linear systems, which is truly significant because most of the real-life input-output relationships are  non-linear in nature \cite{khan2017novel}. The simple architecture of the classical NNs if properly implemented can assure an appropriate performance therefore they are still in use for applications such as: data validation, risk management for the analysis and control of project risks, digit recognition for computerized bank check numbers reading, texture examination to detect microcalcifications on mammograms etc.

Another important application is the fitness track of systems with the help of redundant neural networks, the network helps in monitoring complex manufacturing machines.  
A widely used simple neural network architecture is a Radial Basis Function (RBF) \cite{NL4}. 
It gives desirable results in most of the practical problems. 
RBF neural networks are implemented in face recognition, 
it is also applied in pattern recognition due to the arbitrary nonlinear mapping characteristic 
\cite{khan2018fractional}
. Its application can also be found in the forecasting problems such as the time series forecasting because of the simple topological structure. 
Recently, numerous improvements for RBF architecture have been introduced with the objective to provide better performance. For example, 
an algorithm is proposed to exploit weight structures in RBFNN through Bayesian method, 
another algorithm is proposed to remove noisy samples. This method help in discriminating noisy samples from clean ones using the density information of the sample.  In \cite{khan2016novel}, an adaptive kernel is proposed, the algorithm combines two widely used kernels to improve non-linear mapping of the signal in kernel space. In \cite{FRBF,VPFLMS,RVSSFLMS,RVPFLMS,FCLMS,FLMF,CFLMScomments,mFLMScomments,wahab2019comments}, a fractional order weight update rule is proposed. In \cite{ali2014adaptive}, an adaptive learning method is proposed for system identification task.  In \cite{drmoin}, a novel fusion of multiple kernel is proposed. In \cite{hassan2018kernel}, kernel optimization method is proposed using Nelder Mead Simplex algorithm.

In conventional signal processing approaches signals are treated either in time or in space domain only. Spatio-temporal modeling of systems is becoming a significant area of research in various fields such as health-care, Epidemiology, medical imaging, ecosystem monitoring, business and operations etc. Spatio-temporal analyses provides more advantages over conventional uni-dimensional approaches by harnessing the information from both the temporal and spatial domains. In \cite{StEstimation}, estimation of spatio-temporal neural activity using RBF networks is proposed.  Limited bandwidth and susceptibility to interference limits the use of wireless communication. However, using temporal and spatial processing better trade-offs can be achieved, thus providing better performance. 
In \cite{multiobject}, a multi-object tracking algorithm is proposed with spatial-temporal information and trajectory of confidence. Spatial-temporal correlation has been used to design a model which is more efficient and can deal with missed detection. 
A spatio-temporal neural network is applied for visual speech recognition. 

Motivated by the improvements proposed in the architecture and with the growing field of spatio-temporal based algorithms an extension of RBF neural network is introduced in this paper. Herein, an spatio-temporal variant of RBFNN is proposed for the prediction of chaotic time series. The proposed algorithm utilizes the concept of time-space orthogonality and separately deals with the temporal signal(dynamics) and non-linearity(complexity) of the chaotic series.

The main contributions of our research are:
\begin{enumerate}
	\item A spatio-temporal extension of radial basis function (RBF) neural networks is proposed for Mackey-Glass time series 
	prediction problem.
	\item For the proposed architecture, an stochastic gradient descent-based weight update rule is derived.  
	\item  To show the performance gain achieved by the proposed spatio-temporal RBF-NN (STRBF-NN), hundred independent Monte Carlos simulations were performed and mean results are compared with the standard RBF-NN on similar configurations.
\end{enumerate}
   The rest of the paper is organized as follows, in section \ref{sec:1}, the proposed spatio-temporal RBF neural network is discussed followed by an overview of RBF network architecture and derivation of the weight update rule. Performance is analyzed for time series prediction problem in section \ref{sec:results} and the conclusion of the paper is discussed in section \ref{sec:conclusion}.

\section{The Spatio-Temporal RBF Neural Network} \label{sec:1}
\begin{figure}[h!]
	\begin{center}
		\centering
		\includegraphics[width=8cm]{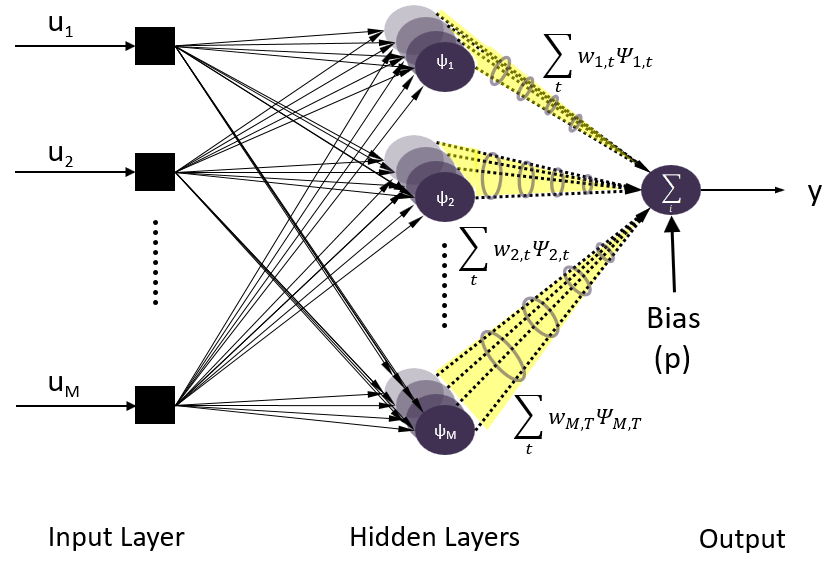}
	\end{center}
	\caption{Signal flow diagram of the spatio-temporal RBF neural network}
	\label{rbfarch}
\end{figure}

Typically the RBFNN consists of 3 layers, including one input and one output layer. The non-linearity in RBFNN comes from a nonlinear hidden layer. However in spatio-temporal processing of signal we also need temporal expansion of signal in kernel space as shown in Fig.\ref{rbfarch}. For conventional architecture, consider input vector $\mathbf{u} \in \mathbb{R}^{S}$ , and the output mapping function, $y:\mathbb{R}^{S}\rightarrow\mathbb{R}^{1}$, is given as
\begin{eqnarray}
y=\sum_{i=1}^{S}w_{i}\psi_{i}(\left\|\mathbf{u}-\mathbf{c}_{i}\right\|)+p
\end{eqnarray} 
where $\mathbf{c}_i \in \mathbb{R}^{S}$ are the centers of the RBF network, $S$ is the number of neurons in the hidden layer, $\mathbf{w}_i$ are the link weights that connect the hidden layer to output neuron, the basis function of the $i$th hidden neuron is represented by the symbol $\psi_i$ and bias $p$ is added in final result of neural network. Generally for easiness only one output neuron is considered.

The output layer of RBF is linear, therefore to make non-linear decision boundary of the features linear. The concept of Cover's theorem \cite{covers} is used, by which mapping from lower dimensional space to higher dimensional make the features space linearly separable in kernel space. Most commonly used mapping functions are as follows \cite{mapping}.\\
Multiquadrics:
\begin{equation}
  \psi_{i}(\|\mathbf{u-c}_{i}\|) = (\|\mathbf{u-c}_{i}\|^{2}+\tau^{2})^{\frac{1}{2}}  
\end{equation}
Inverse multiquadrics:
\begin{equation}
 \psi_{i}(\|\mathbf{u-c}_{i}\|) = \frac{1}{(\|\mathbf{u-c}_{i}\|^{2}+\tau^{2})^{\frac{1}{2}}}
\end{equation}
Gaussian:
\begin{equation}\label{Gradial}
 \psi_{i}(\|\mathbf{u-c}_{i}\|)= \exp \left({\frac{-\|\mathbf{u-c}_{i}\|^{2}}{\sigma_{i}^{2}}} \right)
 \end{equation}
 The spread parameter is given by the symbol $\sigma^{2}$ and $\tau > 0$ is a constant.
\subsection{Gradient descent-based spatio-Temporal RBFNN}
For the training of proposed Spatio-Temporal RBFNN (STRBF-NN), we employed the widely used least square method \cite{abbas2018topological} called gradient descent algorithm. Consider an RBFNN as shown in Fig.\ref{rbfarch}, the final mapping, at the $k$th learning iteration at a particular epoch, is given as:
\begin{eqnarray}
\label{map}
y(k)=\sum_{i=1}^{S}\sum_{t=1}^{T}w_{(i,t)}(k)\psi_{(i,t)}(\mathbf{u},\mathbf{c}_{(i,t)})+p(k)
\end{eqnarray} 
where $T$ is the truncated time, the bias $p(k)$ and synaptic weights $w_{(i,t)}(k)$ are updated at each iteration. The objective to be minimized $\mathcal{O}(k)$:
\begin{equation}
\label{cost}
\mathcal{O}(k)=\frac{1}{2}(d(k)-y(k))^{2} = \frac{1}{2}  e^2(k)
\end{equation}
The target signal at $k$th instant is given by a symbol $d(k)$
whereas $e(k)$ is the immediate difference in estimated target $e(k)=d(k)-y(k)$.

The conventional gradient descent based learning rule is given as:
\begin{equation}\label{eq:WeightConven}
w_{i}(k+1) = w_{i}(k) - \eta \nabla_{w_{i}} \mathcal{O}(k)
\end{equation}

We propose to use the spatio-temporal extension of RBF, thereby equation \eqref{eq:WeightConven} can be given as:
\begin{equation}\label{eq:WeightConven1}
w_{(i,t)}(k+1) = w_{(i,t)}(k) - \eta \nabla_{w_{(i,t)}} \mathcal{O}(k)
\end{equation}
where $\eta$ is the step size, $w_{(i,t)}(k)$ and $w_{(i,t)}(k+1)$ are the current and updated weights.  
Evaluating the factor $-\nabla_{w_{(i,t)},}\mathcal{O}(k)$ by making use of the chain rule: 
\begin{equation}\label{Normal} \nabla_{w_{(i,t)}} \mathcal{O}(k) =   \frac{\partial \mathcal{O}(k)}{\partial e(k)} \times\frac{\partial e(k)}{\partial y(k)} \times \frac{\partial y(k)}{\partial w_{(i,t)}(k)} 
\end{equation}
After taking partial derivatives equation (\ref{Normal}) is simplified to:
\begin{equation}\label{Normal_Sim}
 \nabla_{w_{(i,t)}} \mathcal{O}(k) = -  \psi_{(i,t)}(\mathbf{u},\mathbf{c}_{(i,t)}) e(k)
\end{equation}

using \eqref{Normal_Sim} equation \eqref{eq:WeightConven1} is turn out to be:		
\begin{equation}\label{weightf}
w_{(i,t)}(k+1) = w_{(i,t)}(k) + \eta  \psi_{(i,t)}(\mathbf{u},\mathbf{c}_{(i,t)}) e(k) 
\end{equation}

In the same way the learning rule for $p(k)$ is given as:
\begin{equation}
p(k+1) = p(k) +  \eta e(k) 
\end{equation}

For complete weight vector $\mathbf{w}$, we can write Eq \eqref{weightf} as
\begin{equation}
\mathbf{w}(k+1) = \mathbf{w}(k) + \eta \; \mathbf{\psi}(\mathbf{u},\mathbf{c}) e(k)
\end{equation}

\section{Experiments Results}\label{sec:results}
For the experimental setup, the chaotic time series prediction problem is considered.  In particular, Mackey-Glass series is used.  The Mackey-Glass time series is represented by a delayed differential equation given by \cite{GlassSeries}
\begin{equation}
	\label{Mackey_Glass}
	\frac{du(t)}{dt}=\frac{au(t-\tau)}{1+u(t-\tau)^{10}}-bu(t)
\end{equation}	 
Here, $\tau = 20$, $a = 0.2$, $b = 0.1$, and
$u(t-\tau) = 0$, for $(\tau \geq t \geq 0)$. Therefore, a series $u(t)$ for
$t = 1, 2, . . . ,3000$, is derived by Eq. (\ref{Mackey_Glass}).

\begin{figure}[h!]
	\begin{center}
		\centering
		\includegraphics*[width=8cm]{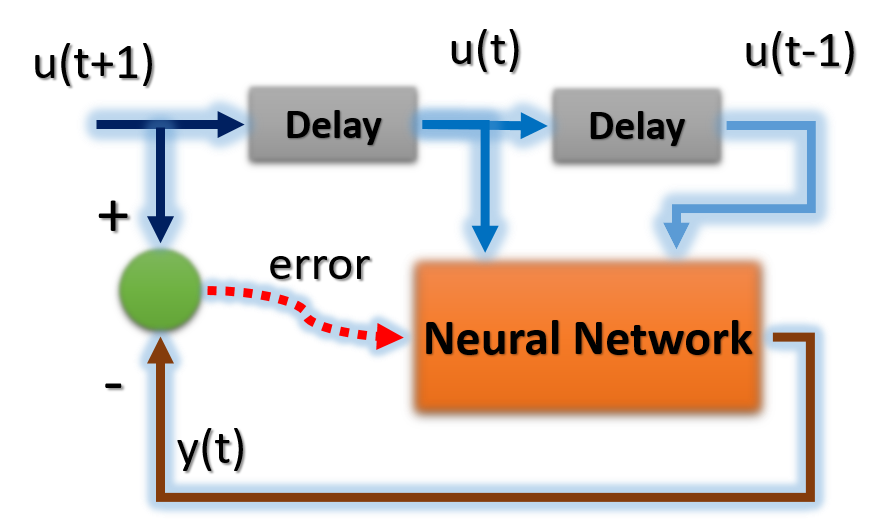}
	\end{center}
	\caption{Typical time series prediction model}
	\label{fig: Time_Seris_Model}
\end{figure}

The time series data is produced by performing sampling of the curve $u(t)$ at one second intervals.  Moreover, due to the addition of white Gaussian noise, the training dataset achieves an SNR of $30$dB. For simulation of both the RBF configurations, training samples from $100 \leq t \leq 2500$ is used, while for validation purpose, samples in the range $2500 < t \leq 3000$ were used. Both the RBF and STRBF are trained with two samples (one current value and one previous instance value) as input and a very next future value as an target. The cursor frame of input vector moves forward in an overlapping manner as depicted in Fig. \ref{fig: Time_Seris_Model}. The simulation is iterated for hundred Monte-Carlos runs, and the average of the results obtained during the experiments are reported. 
For each iteration bias (p) and weights (w) were initialized by random numbers.

\begin{figure}[h!]
	\begin{center}
		\centering
		\includegraphics*[width=9cm]{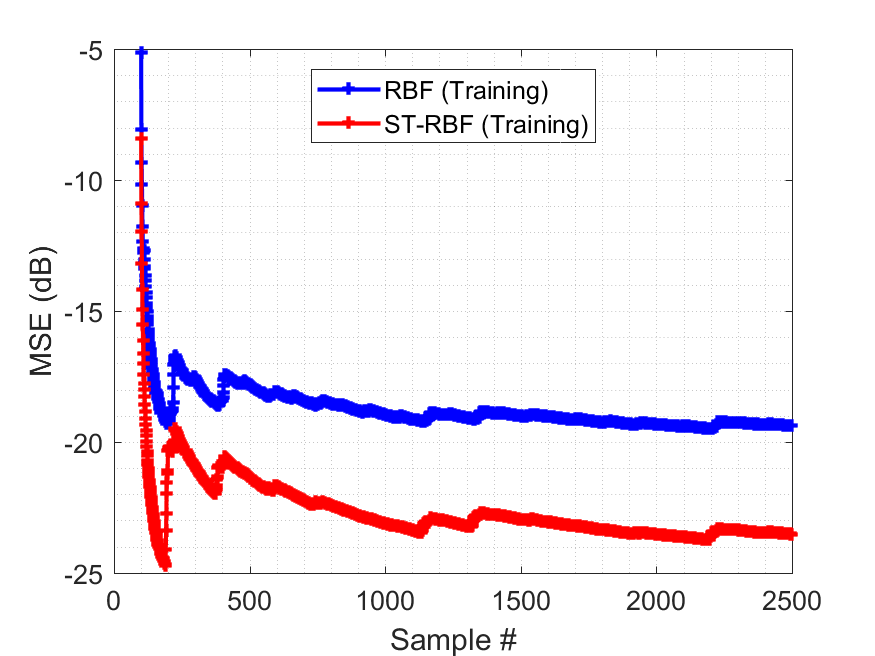}
	\end{center}
	\caption{Training MSE curves of the time series prediction problem}
	\label{TSMSE}
\end{figure}

The hidden layer of conventional RBF contains $20$ neurons in spatial axis in order to map the non-linearity of the signal in higher dimensions, and the end-node linear layer contains just only a single neuron, while the proposed STRBF contains two parallel temporal layers of $10$ neurons in each to map the temporal dynamics and non-linearity of the signal. Gaussian kernel function is used for all the neurons, whereas for the output neuron, a linear summation function is used. The input layer is of size $2$. Fig. \ref{fig: Time_Seris_Model} displays a illustration of time series prediction system. In RBF learning phase, the step-size of gradient-decent $\eta$ was set to be $1\times10^{-2}$, whereas, learning rate $\eta$ of STRBF is set to $5\times10^{-2}$. The variance and mean of every Gaussian kernel is derived using K-means clustering algorithm. In order to achieve the same rate of convergence, the learning rate and standard deviation ($\sigma$) of the Gaussian kernel for the proposed STRBF is kept half of that of conventional RBF. The training MSE graphs of the conventional RBF network and the proposed STRBF network are displayed in Fig \ref{TSMSE}.


After the completion of training, STRBF-NN attained an average mean squared error (MSE) of $-23.52$dB. The RBF-NN shows a relatively greater MSE of $-19.38$dB, despite using the same number of neurons in both the configurations.

\begin{figure}[h!]
	\begin{center}
		\centering
		\includegraphics*[width=9cm]{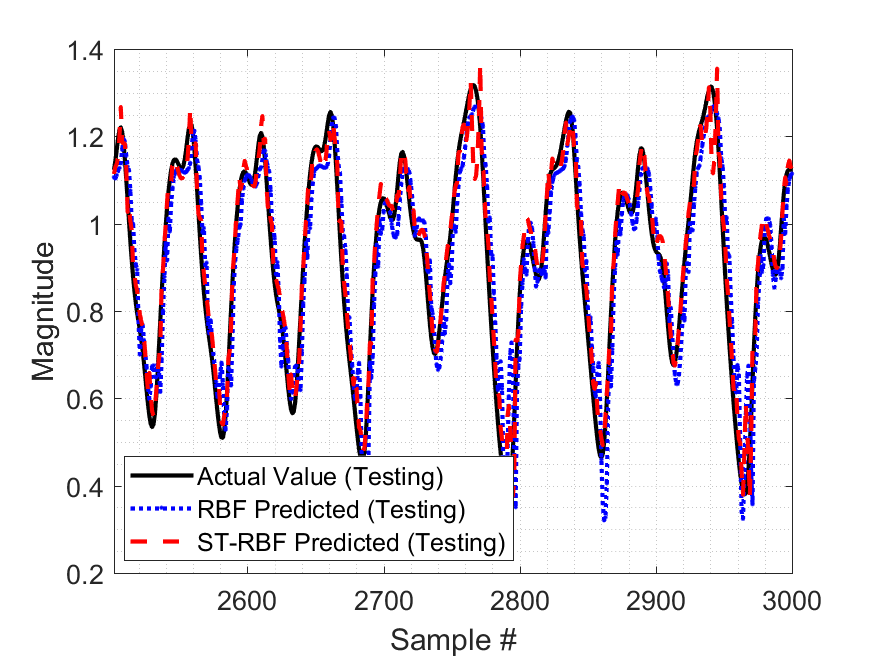}
	\end{center}
	\caption{Testing data (actual vs. predicted) of STRBF and conventional RBF}
	\label{TSOUT_Test}
\end{figure}

\begin{figure}[h!]
	\begin{center}
		\centering
		\includegraphics*[width=9cm]{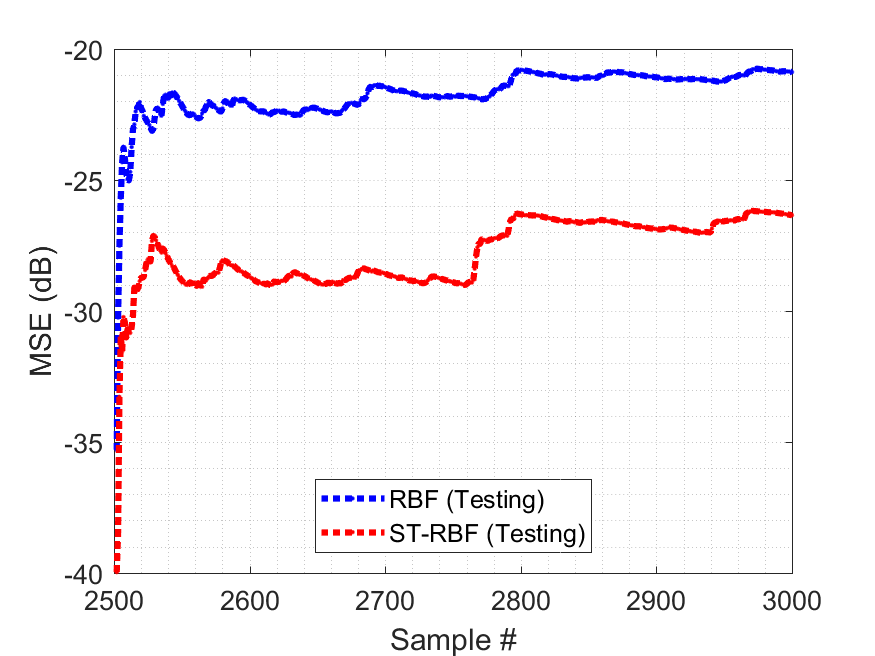}
	\end{center}
	\caption{Testing MSE curves}
	\label{TSMSE_Test}
\end{figure}

Both the RBF configurations were simulated on unseen samples in the testing phase. In fig \ref{TSOUT_Test}, the comparison of predicted data of both the configurations and the actual data is displayed. The proposed STRBF network predicted the data points that are much closer to the actual data samples.  For the quantitative measure, in Fig \ref{TSMSE_Test}, we presented the MSE curves. The STRBF-NN outperforms conventional RBF-NN by attaining a mean MSE of $-26.34$dB compared to the RBF-NN which achieved $-20.88$dB MSE. The results of the prediction problem are enlisted in Table~\ref{TS:CORR}.

\begin{table}[!h]
	\begin{center}
		\caption{Statistics Comparison}		\label{TS:CORR}
		\begin{tabular}{|c|c|c|}
			\hline  \textbf{Configuration} & \textbf{Learning Phase} & \textbf{Validation} \\
			& \textbf{mean square error (dB)} & \textbf{mean square error (dB)}\\ 
			\hline  RBF-NN & -19.38 & -20.88 \\
			\hline  \textbf{STRBF-NN} & \textbf{-23.52} & \textbf{-26.34} \\ 
			\hline 
		\end{tabular}
	\end{center}
\end{table}

\section{Conclusion}\label{sec:conclusion}
In this paper an spatio-temporal extension of RBF neural network is proposed for the prediction of chaotic time-series. In particular, two step forward signal of Mackey-Glass time-series is predicted.  The proposed algorithm is compared with standard RBF-NN. Gradient descent-based learning algorithm is used for the training of adaptive weights parameters. The MSE learning curves showed the overall best results of the proposed STRBF-NN in comparison to the conventional RBFNN for $100$ independent runs. Here we would like to point out that the proposed spatio-temporal RBFNN is an improvement is the architecture of standard RBF, its performance can be enhanced by incorporating better stochastic learning methods such as q-gradient descent \cite{sadiq2019enhanced,qLMF}, further improvements can be made by adaptive learning of the optimal network connections, or through the introduction of dropouts probabilities etc. Code and supplementary material is available online \cite{STRBF_TS_CODE}.
	

\end{document}